\documentclass[conference]{IEEEtran}
\IEEEoverridecommandlockouts
\usepackage{cite}
\usepackage{amsmath,amssymb,amsfonts}
\usepackage{algorithmic}
\usepackage{graphicx}
\usepackage{subfigure}
\usepackage{textcomp}
\usepackage{xcolor}
\usepackage{array}
\usepackage{booktabs}
\usepackage{caption}
\def\BibTeX{{\rm B\kern-.05em{\sc i\kern-.025em b}\kern-.08em
    T\kern-.1667em\lower.7ex\hbox{E}\kern-.125emX}}
\begin{document}

\title{Unsupervised Spiking Neural Network Model of Prefrontal Cortex to study Task Switching with Synaptic deficiency\\
}
\author{\IEEEauthorblockN{K.Ashwin Viswanathan}
	\IEEEauthorblockA{\textit{Computer Science Department} \\
		\textit{Oklahoma State University}\\
		Stillwater, USA \\
		ashwin.kannan@okstate.edu \\}
	\and
	\IEEEauthorblockN{Goutam Mylavarapu}
	\IEEEauthorblockA{\textit{Department of Computer Science} \\
		\textit{University of Central Oklahoma}\\
		Edmond, USA \\smylavarapu@uco.edu}\\

	\and\IEEEauthorblockN{Johnson P Thomas}
	\IEEEauthorblockA{\textit{Computer Science Department} \\
		\textit{Oklahoma State University}\\
		Stillwater, USA \\
		johnson.thomas@okstate.edu}\\
	\and

}

\maketitle

\begin{abstract}
In this study, we build a computational model of Prefrontal Cortex (PFC) using Spiking Neural Networks (SNN) to understand how neurons adapt and respond to tasks switched under short and longer duration of stimulus changes. We also explore behavioral deficits arising out of the PFC lesions by simulating lesioned states in our Spiking architecture model. Although there are some computational models of the PFC, SNN's have not been used to model them. In this study, we use SNN's having parameters close to biologically plausible values and train the model using unsupervised Spike Timing Dependent Plasticity (STDP) learning rule. Our model is based on connectionist architectures and exhibits neural phenomena like sustained activity which helps in generating short-term or working memory. We use these features to simulate lesions by deactivating synaptic pathways and record the weight adjustments of learned patterns and capture the accuracy of learning tasks in such conditions. All our experiments are trained and recorded using a real-world Fashion MNIST (FMNIST) dataset and through this work, we bridge the gap between bio-realistic models and those that perform well in pattern recognition tasks.

\end{abstract}

\begin{IEEEkeywords}
unsupervised, pattern recognition, neural network, artifical neural network, computational intelligence, bio inspired
\end{IEEEkeywords}

\section{Introduction}
Decision making is one of the fundamental cognitive processes of human behaviors. Evidence arising out of cognitive studies show the PFC region of the brain is involved in decision making and task switching activities \cite{braver1995computational,rolls2008computational}. There has been a renewed interest in SNN as they imitate biological neurons and have been used in a variety of supervised and unsupervised tasks \cite{diehl2015unsupervised}. In our previous study of task switching \cite{viswanathan2020study}, we used a computationally tractable model of SNN to implement a bio-inspired neuron model to analyze the effects of task switching from the perspective of the synaptic weight. In this study, we implement a more realistic architecture of the PFC based on bio-realistic connectionist models \cite{braver1995computational,rolls2008computational} and study the behavioral deficits and learning impairment suffered by patients with the PFC lesions within the context of task switching. We explore these avenues using a Hebbian-based biological unsupervised learning rule known as STDP \cite{gupta2009hebbian,song2000competitive}. Our contributions in this paper are: (a) Implementing a biologically realistic architecture of the PFC neurons using SNN, (b) Studying behavioral and memory deficits by simulating lesions in our SNN model, (c) Using a real-world dataset to mimic real-world possibilities in learning and memory formation (d) Using a computationally efficient and tractable framework of SNN to record and graphically plot neural phenomenon like synaptic lesions and memory impairment when learning patterns.Finally, unlike previous works which take a top-down approach where human participants engage in task switching experiments based on cues or use traditional artificial neural networks, we employ a bottom-up approach at a biologically realistic neuronal level. We take this approach so as to understand the detailed biological neuronal mechanisms of task switching. This paper is organized as follows: In Section \ref{rel_work}, we provide a background for this work. Section \ref{bio_frame} talks about the neuron model design, learning, dataset used and our the PFC architecture. In Section \ref{exp} we present the experiments. This study concludes in Section \ref{res} where we present and analyze the results.

\section{Related Work}\label{rel_work}
Studies like \cite{grant1948behavioral,rubinstein1994task} support the evidence of the PFC as the region most involved in enabling decision making and task switching in the brain. In these works, the PFC decision-making is tested utilizing task switching tasks performed by human participants. Moreover, these experiments take place in a controlled manner by providing cues to participants before enabling changes in tasks. In our model, we address this issue by training in an unsupervised manner using STDP and analyze the response of neurons. Although there have been studies on task switching using computational models \cite{braver1995computational,lynch2019winner,rolls2008computational}, they use traditional neural networks and train the models in a supervised manner. Our model improves upon these studies by using a spiking neural architecture which provides for more realistic biological behavior. We use the Leaky Integrate and Fire (LIF) type of neurons to simulate the spiking behavior. Lesions and behavioral deficits arising out of these are explored in works like \cite{rolls2008computational,cohen1996computational} which also uses task switching to gauge the memory learning capabilities of the PFC. These studies provided the validation for our results. Our model also implements sustained or persistent neural activity which acts as short-term or working memory. This is used to retrieve an entire episode of memory from a partial stimulus. We validate our model results arising from sustained activity by comparing it with experimental results and studies done in works like \cite{rolls2008computational,rolls2007memory,hertz2018introduction}. Previous research like  \cite{rogers1995costs,braver1995computational} specify a short and longer time duration for experiments on human participants. We also use the same time duration shifts for testing our models with lesions by deactivating synapses. 

\section{Methods}\label{bio_frame}
\subsection{Neuron Design Model}
SNN's mimic real neurons due to their behavior of firing spikes. In our research, we implement the LIF model due to its computational tractability \cite{paugam2012computing}. They are widely used to simulate computational models as they capture a variety of neuronal behavior dynamics \cite{gerstner2002spiking}. Let the membrane potential of LIF neurons be represented by $V_{m}$. We can then define our LIF neurons in terms of Ordinary Differential Equations to describe the evolution of $V_{m}$ over time. We can define our synaptic equations as follows:
\begin{equation}\label{syn_leak}
	I_{leak}=g_{leak} * (E_{leak} - V_{m})
\end{equation}
\begin{equation}\label{syn_exc}
	I_{exc}=g_{exc} * (E_{exc} - V_{m})
\end{equation}
\begin{equation}\label{syn_inh}
	I_{inh}=g_{inh} * (E_{inh} - V_{m})
\end{equation}
Here $g_{leak}$ refers to the leak conductance and stays constant and is stored in leak current $I_{leak}$. $g_{exc}$ is the excitatory conductance and reflects the excitatory input $I_{exc}$. Similarly $g_{inh}$ is the inhibitory conductance and measures the strength of inhibitory input $I_{inh}$. $E_{leak}$, $E_{exc}$ and $E_{inh}$ are the leak, excitatory and inhbitory membrane potentials. Combining (\ref{syn_leak}), (\ref{syn_exc}) and (\ref{syn_inh}) the membrane equation with several synapses is given as:
\begin{equation}\label{mem_eq}
	C_{m}\frac{dV}{dt}=I_{leak} + I_{exc} + I_{inh} + \xi
\end{equation}
where $C_{m}$ is the membrance capacitance constant specified in Farads and $\xi$ is the standard noise term. Evolution of the inhibitory and excitatory conductance over time is given by


\begin{equation}
	\frac{d{g_{inh}}}{dt}=-\frac{g_{inh}}{\tau_{inh}}
\end{equation}
\begin{equation}
	\frac{d{g_{exc}}}{dt}=-\frac{g_{exc}}{\tau_{exc}}
\end{equation}
where $\tau_{inh}$ and $\tau_{exc}$ are the membrane time constants. Parameters used for neurons are close to actual biological values in all our experimental trials. Spiking Neurons communicate by generating and propagating electrical impulses known as Action Potentials or Spikes. After firing a spike the membrane potential $V_{m}(t)$ is reset to the resting potential $V_{rest}$. During this phase, the neuron goes to a refractory period $\tau_{ref}$ and cannot spike again during this phase. We can model the spiking behavior of LIF neurons having spike voltage threshold $V_{\phi}$ by:

\renewcommand{\arraystretch}{2.2}
\begin{table}[t]
	\caption{Spiking Neural Network parameters.}
	\label{tab:SNN_tbl}
\begin{center}

	\begin{tabular}{|p{4cm}|p{4cm}|}
		\hline
		\textbf{Parameter} & \textbf{Value} \\ \hline
		$V_{\phi}$ &  {$-55mV$} \\ \hline
		$V_{rest}$ &  {$-70mV$} \\ \hline
		$E_{inh}$ & {$-75mV$}\\ \hline
		$E_{exc}$ & {$0mV$} \\ \hline
		$E_{leak}$ & {$-65mV$} \\ \hline
		
	\end{tabular}
\end{center}
\end{table}

\begin{equation}\label{spike_eq}
	V_{m}(t)=
	\begin{cases}
		V_{reset}, & \text{if }
		\!\begin{aligned}[t]
			V_{m}(t)& > V_{\phi}\\
		\end{aligned}
		\\
		V_{m}(t), & \text{otherwise}
	\end{cases}
\end{equation}
The various biological parameter values are specified in Table \ref{tab:SNN_tbl}

\begin{figure*}[t]
	\centering
	\captionbox[Caption]{Spiking Neural Architecture of PFC .\label{fig:SNN_arch}}{
		\includegraphics[scale=0.55]{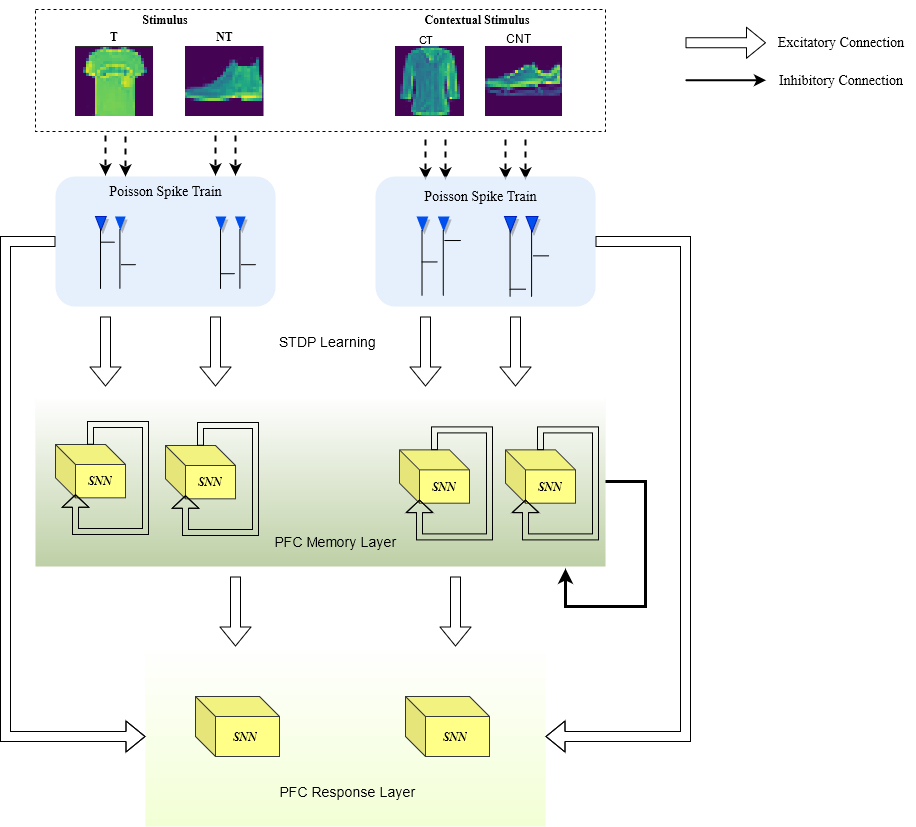}
	}
	\label{SNN_arch}
\end{figure*}

\subsection{Unsupervised Plasticity Learning Rule}
In our study, we use the unsupervised synaptic plasticity learning rule known as Spike Timing Dependent Plasticity (STDP). STDP has been linked to various learning rules and mechanisms like Hebbian Learning and short-term prediction\cite{rao2002predictive,mehta2001neuronal} of biological neurons. This type of biologically-inspired learning rule is used in SNN. Learning of patterns takes place by modification of synaptic weights in response to the firing times of pre and post-synaptic spikes. This leads to observing two neural phenomena namely, Long Term Potentiation (LTP) \cite{gupta2009hebbian} which occurs when the neurons are learning patterns and Long Term Depression (LTD) \cite{gupta2009hebbian,bi1998synaptic} which leads to forgetting of patterns in the network. As STDP depends on the timing of pre and post-synaptic spikes, we can define the STDP function by \cite{bi1998synaptic} where $w$ is the synaptic weight:
\begin{equation}
	(\Delta w) = \begin{cases}
		A^+ e^{-\Delta t/\tau^+} & \Delta t>0 \\
		A^- e^{\Delta t/\tau^-} & \Delta t<0
\end{cases}\end{equation}
where $A^+$ holds the synaptic trace resulting from LTP and $A^-$ stores the LTD synaptic trace. $\tau^+$ and $\tau^-$ denote the change in pre and post-synaptic spike time. $\Delta t$ is the time delay difference of pre-synaptic and post-synaptic spikes given by $\Delta t=\tau^- -\tau^+$.  In case of LTP, the synaptic weight $w$ is updated by:
\begin{align}\label{eq:pre_eq}
	\begin{split}
		a^- &\rightarrow a^- + A^-\\
		w &\rightarrow w+a^+
	\end{split}
\end{align}
Updating the weight when there is depression in the network is given by:
\begin{align}\label{eq:post_eq}
	\begin{split}
		a^+ &\rightarrow a^+ + A^+\\
		w &\rightarrow w+a^-
	\end{split}
\end{align}
For computational efficiency, we store only the synaptic traces and the weight $w$ is bound by $0 <= w<=w_{max}$.

\subsection{Unsupervised Spiking Neural Architecture}
Our architecture draws on aspects derived from computational neuroscience and psychology \cite{rolls2008computational}. Our network is depicted in Figure \ref{fig:SNN_arch}. It consists of three layers namely the Input Sensory Layer which receives the input stimuli and encodes them into Poisson spike trains and transfers the encoded spikes to both the PFC Memory layer and the PFC Response layer. Neuron and STDP parameters are set to biologically realistic values. In the following sections, we will describe the role and functions of these layers.

\subsubsection{Input Sensory Layer}
This is the input layer. We convert the input stimuli into spike trains having frequency proportional to the value of each pixel. This type of encoding is known as Rate coding. Within the cortex, neuronal firing activity is irregular and stochastic \cite{moreno2014poisson}. This behavior is captured by a Poisson process and spiking activity in the brain roughly approximates a Poisson distribution \cite{masquelier2008spike}. Hence we overlay the input spike train as a Poisson process. The input layer is made up of Excitatory neurons which are mapped to each pixel in the input row. SNN's carry information in spikes which are transmitted between different layers via Synapses.
\begin{figure*}[t]
	\centering
	\captionbox[Caption]{Sustained Activity in the Memory Layer.\label{fig:SNN_sus_act}}{
		\includegraphics[scale=0.62]{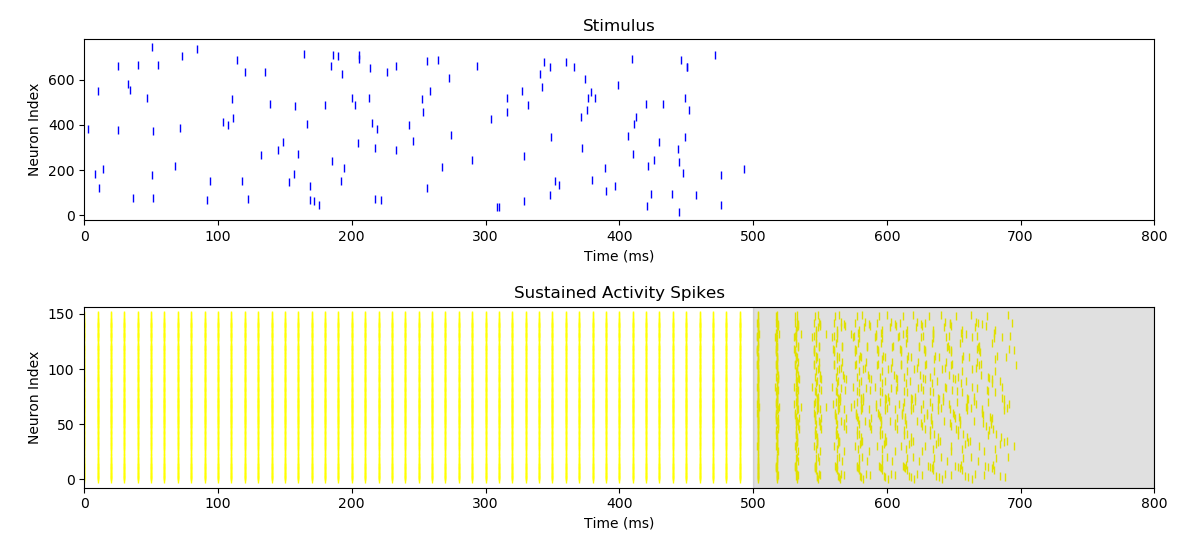}
	}
	\label{SNN_sus_act}
\end{figure*}
\subsubsection{PFC Memory Layer}
This layer receives the stimulus through excitatory synapses from the Input Sensory Layer and makes lateral self-inhibitory synaptic connections to generate competition among neurons. This mode of lateral self-inhibition is known as Winner Take All. This ensures a single firing neuron is picked which fires consistently amongst competing neurons \cite{lynch2019winner,gupta2009hebbian}. Weight matrices of the neurons act as the receptive fields and they are either strengthened or weakened based on the timing of spikes arriving from the Input layer. STDP plasticity rule enables the strengthening of synapses based on the response of neurons to stimuli \cite{song2000competitive}.
\subsubsection{Sustained Activity} From our model in Figure \ref{fig:SNN_arch} we observe that each LIF unit in the Memory layer has self excitatory connections. This helps the PFC neurons to carry a trail of learned activity from the input stimulus and maintain them for a short duration in the absence of any input. This type of behavior dynamics is known as an attractor network \cite{rolls2008computational}. This persistent firing activity in the absence of any input or stimulus is known as Sustained Activity. This phenomenon is responsible for the formation of short-term or working memory \cite{rolls2007memory}. These representations or knowledge are then maintained in a stable manner even after ceasing to activate the network using any input stimuli.

Several research studies have shown that persistent neuronal activity has been observed whenever the PFC is involved in decision making or selection of tasks among other activities \cite{goard2016distinct,riley2016role,rainer1999prospective}. Figure \ref{fig:SNN_sus_act} illustrates persistent spiking activity in our architecture modeled using LIF neurons. As seen in Figure \ref{fig:SNN_sus_act}, we present the Target (T) stimulus which is a $28 * 28$ matrix of image arrays from FMNIST dataset for a duration of $500ms$ and is stopped. The neurons continue spiking beyond $500ms$ up to $800ms$ indicated by the shaded portion and gradually decline as the inhibitory neurons suppress the activity of excitatory neurons which is one of the key functions of the PFC \cite{rolls2007memory}. This continued spiking by neurons holds knowledge of the learned pattern for a short duration which can be used to reconstruct the entire pattern from the whole memory when tested with contextually similar inputs to the original input.

\subsubsection{Response Layer}
This is the decision-making layer and the responses are recorded by observing which of the two units of LIF neurons spike first on seeing a target stimulus. This layer receives excitatory signals from  the Input and Memory layer. The target and non-target neuron is denoted by $N_{t}$ and $N_{nt}$. Recordings are made by noting which of the neurons membrane potential $V_m$ first exceeds the threshold $V_{\phi}$. 

\begin{figure*}[t]
	\centering
	\captionbox[Caption]{Neuron responses with full Synapse connectivity under short and long delay between task switching.\label{fig:SNN_result_full}}{
		\includegraphics[scale=0.85]{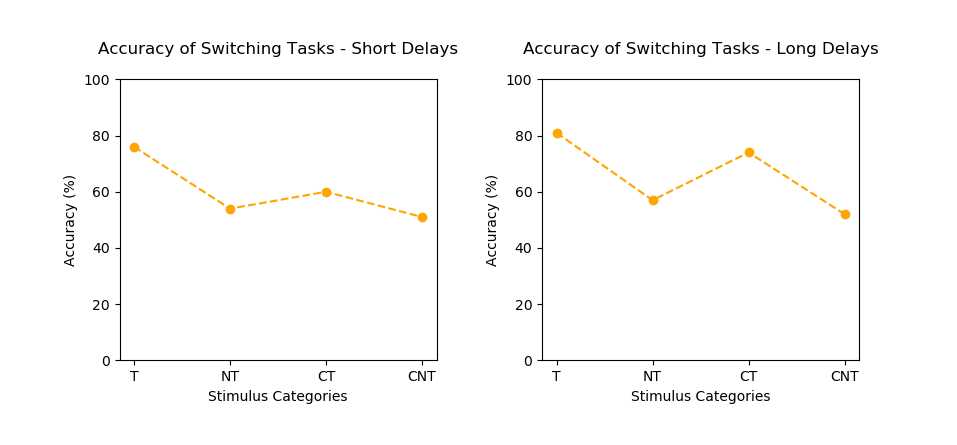}
	}
	\label{SNN_result_full}
\end{figure*}
\begin{figure*}[h]
	
	\centering
	\captionbox[Caption]{Final STDP weights for Target Stimulus at full synaptic connectivity. \label{fig:SNN_wt_full}}{
		\includegraphics[scale=0.85]{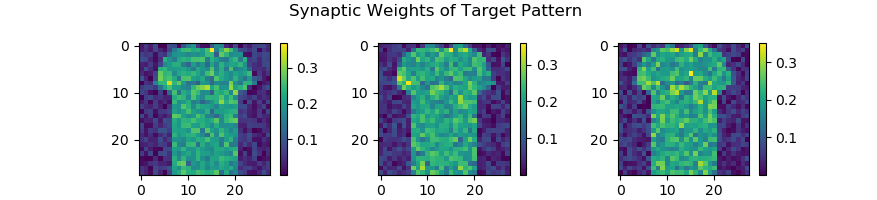}
	}
	\label{SNN_wt_full}
\end{figure*}

\section{Experiments}\label{exp}
We validate our architecture by devising experiments to observe the learning and adaptive properties of the PFC neurons. We use a real-world Fashion-MNIST (FMNIST) image dataset which is suitable for classification and image recognition tasks. Our goals in these experiments are the following: (a) Simulate experiments based on previous research studies in the areas of cognitive psychology \cite{rogers1995costs} and brain frontal cortex lesions \cite{cohen1996computational,reggia1999disorders}, (b) Analyze and record the learning and adaptation behavior of the PFC neurons when stimuli are switched at longer and shorter time durations, (c) Observe the memory formation capacity when synaptic connections between the Input and Memory module are switched off partially. This simulates a network having neuronal lesions \cite{rolls2008computational,reggia1999disorders}. FMNIST dataset comprises $70,000$ records with $10$ categories of fashion products. The labels consist of $7000$ images each, with $60000$ training and $10000$ test samples. Every row represents a $28 * 28$ shaped matrix. For this study, we chose four stimulus types (i.e.) Target, Non-target, Context-Target and Context-Non-target. The model learns patterns by modifying the synaptic weight $w$ of neurons $N$ and is represented as $w: N  \times N$ describing the synaptic connections between neurons in the network. Weight $w$ gains higher values whenever a post-synaptic spike is preceded by pre-synaptic spike (i.e.) $\tau_{pre} < \tau_{post}$. For every pixel $p_{1}, p_{2},\dots, p_{i}$ where $i \in 1, 2, 3, \dots, 784$ correspond to the number of features, we map each pixel $p_{i} \rightarrow N_{exc}^i$ to excitatory neurons. The input and output are excitatory such that $N_{input} \cup N_{output} \subseteq N_{exc}$. The \textit{Target} is used predominantly and serves as the stimulus responsible for eliciting a spike from the target neuron $N_{t}$. In the FMNIST dataset, this image falls under the class of \textit{t-shirts}. \textit{Non-Target} should not evoke a spike response from $N_{t}$ and its effect on the model is observed in the dynamics of output neuron $N_{nt}$. This image comes under the class of \textit{ankle boots}. Target and Non-Target stimuli are presented in a $70:30$ mix with the target signals appearing in the majority of our trials. As the PFC model exhibits persistent neural activity, we use two additional sets of stimuli to test this functionality. The purpose of \textit{Context-Target} and \textit{Context-Non-Target} is used to analyze the pattern recognition capacity of the short-term working memory of the PFC neurons. These two stimuli are presented after the model learns patterns from target and non-target. This leads to sustained activity where the neurons maintain a steady firing state despite the absence of input. In this state we present context-target and context-non-target to record how the neurons use this short-term memory to retrieve the whole pattern. Capturing the firing activity of $N_{t}$ and $N_{nt}$ when they are in their persistent state helps in better understanding of the PFC decision-making capabilities. To effectively test this, we choose stimulus contextually similar to the main target and non-target images. We used cosine similarity to get images similar to Target and Non-Target. An image having high similarity value to the target is chosen as our context-target and similarly, we choose an image for context-non-target. The $4$ types of input stimuli are shown in Figure \ref{fig:SNN_arch}. For studying the effects of lesions, we partially deactivate  synapses between the Input Sensory and the PFC Memory layer.

\section{Results}\label{res}

\begin{figure*}[t]
	\centering
	\captionbox[Caption]{Neuron responses with Deactivated Synapse connectivity under short and long delay between task switching. \label{fig:SNN_result_deac}}{
		\includegraphics[scale=0.85]{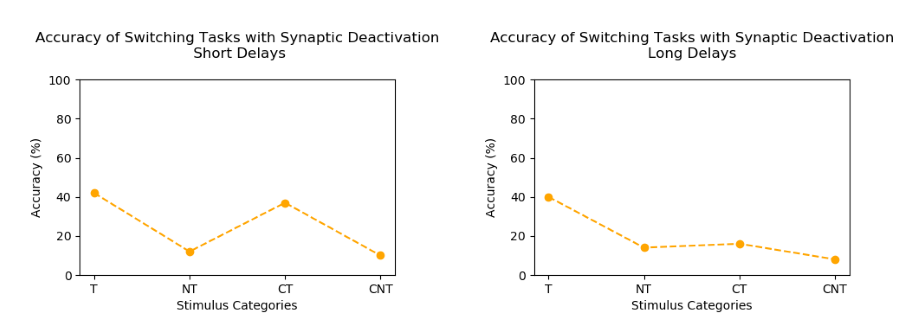}
	}
	\label{SNN_result_deac}
\end{figure*}
\begin{figure*}[h]
	\centering
	\captionbox[Caption]{Final STDP weights for Target Stimulus with Partial synaptic connectivity. \label{fig:SNN_wt_part}}{
		\includegraphics[scale=0.85]{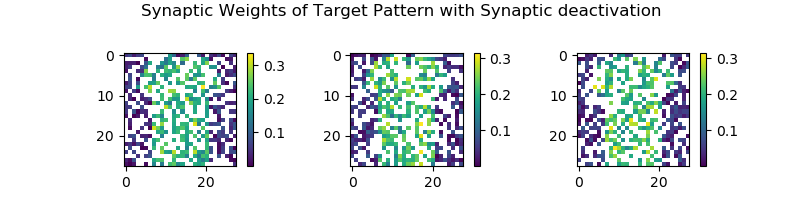}
	}
	\label{SNN_wt_part}
\end{figure*}

Our goal is to capture and understand how the neurons in our PFC model change to different stimulus conditions having full and partial synaptic connectivity. This aids in understanding of how memory and learning get impaired in the absence of synapses between neurons. We discuss our results obtained for the different trials of experiments conducted. The responses for full synaptic connection between all the layers in our model is illustrated in Figure \ref{fig:SNN_result_full}. We performed $1000$ trials for every stimulus category and recorded the responses. From Figure \ref{fig:SNN_result_full}, the model responds well to the target stimulus with an accuracy of $76\%$. As STDP is unsupervised, there was no prior training of the model and this simulates real-world conditions of task switching on demand. The learning of patterns in this trial can be attributed to LTP which causes the synaptic weight $w$ to increase \cite{gupta2009hebbian}. In task-switching with short delays, we switch task stimulus every $350ms$. For the non-target stimulus, the model performs at $54\%$ accuracy indicating adjustment to the new task leading to the synaptic weights undergoing depression \cite{viswanathan2020study}. We then analyze the effect of short-term memory by presenting the context-target and observe the neurons responding with $60\%$ correct firing of spikes. This higher accuracy is because of dense excitatory connections within the memory module layer as shown in Figure \ref{fig:SNN_arch}. These recurrent excitatory connections provide a higher firing activity which is then regulated by the local self-inhibition in the memory layer. This enables the neurons to maintain their firing activity for a short duration and are able to retrieve an entire memory from a partial stimulus. This phenomenon occurs in real neurons of the PFC as discussed in previous research works \cite{rolls2008computational,rolls2007memory,hertz2018introduction}. We observe a similar higher accuracy for context-non-target due to the effect of sustained firing activity. Similar to our previous study on Task Switching using SNN's \cite{viswanathan2020study}, we notice that on longer durations, the model is better able to adapt to the incoming patterns with higher accuracy of $81\%$ for the target. Here, the switch occurs every $550ms$. The higher accuracy is attributed to the  neurons storing sufficient information in their memory and are better able to adapt when tasks are switched \cite{sudevan1987cuing}. For context-target stimulus the accuracy is $74\%$ and $57\%$ for non-target and $52\%$ for context-non-target tasks. These results show that our network can reproduce most of the behavioral responses as found in the following study \cite{braver1995computational}, which performed similar task-switching trials on human participants. We show the final synaptic weight representation for the target stimulus learned by the model using STDP in Figure \ref{fig:SNN_wt_full}. We now simulate the network by deactivating synapses between the Input and Memory layer. This causes deficiency in the information being transmitted and is analogous to a lesion in the cortical regions of the brain \cite{reggia1999disorders}. Altering the synaptic connectivity inhibits the model from learning representations thereby impairing the decision-making capability of PFC. This leads to an absence of short-term or working memory \cite{durstewitz2000dopamine} leading to poor learning and decision making. We shift our focus to study the response accuracy of neurons for partial deactivation of synapses. We deactivate the synapse connections between the Input sensory layer and the Memory layer by a probability value of $Syn_{p} = 0.5$. This gives the model a $50\%$ chance of making a successful synaptic connection and transmitting spike trains. From Figure \ref{fig:SNN_result_deac}, for short delays, the accuracy response to target stimulus is $42\%$ which is low compared to the accuracy with full synapse connectivity. This shows that patients decision-making and memory formation are impaired in lesioned neuorns as the dopamine needed to enable sustained activity is deficient \cite{durstewitz2002computational}.
Accuracy for context-target is $37\%$ and comes close to target (T) accuracy due to the presence of partial representations held actively by weak persistent neurons in a deficient synaptic network. This scenario showcases our model's ability to learn patterns with partial synaptic connectivity between its layers which happens due to STDP. Non-target accuracy is $12\%$ and $10\%$ for context-non-target. For longer delays with synaptic deficiency, the accuracy for target stimulus is $40\%$, but there is a rapid fall in response for the context-target giving an accuracy of only $16\%$. This happens as PFC lesions have been shown to have more impact when there is a longer delay between task switches \cite{stuss1984neuropsychological,rubinstein1994task}. There is a $350ms$ gap between stimulus presentation in longer delays and during this interval of time, the weakened sustained activity by the neurons decays leading to little or no working memory. Accuracies for non-target and context-non-target are even lower as they do not occur frequently. Lesions in synaptic activity cause the neurons to have partial information with no spiking activity in some neurons. This causes higher error rates in learning leading to a high cost of switching \cite{shallice2008multiple,rubinstein1994task}. This leads to incorrect firing activity leading to depression which is characterized by $\tau_{post} < \tau_{pre}$ and brings down the accuracy of correct responses. The impartial pattern representations formed by lesioned synaptic weights $w$ by STDP is seen in Figure \ref{fig:SNN_wt_part} which shows our model is capturing the general pattern of the image but deficient synaptic connectivity prevents it from learning as a whole.

\section{Conclusions}
We have presented a computational model of the PFC using Spiking LIF neurons to study the mechanism of learning and decision making using task-switching trials. The results of our model align with experiments performed on human and primate subjects \cite{yeung2003effects,funahashi1989mnemonic,rubinstein1994task,braver1995computational}. They also follow the biological process observed in real neurons having lesions. We showcase neural phenomenon like LTP and LTD which occurs in real neurons during the formation of memory. Additionally, our model also exhibits persistent attractor network states due to recurrent self excitatory connections. These connection dynamics have their basis in the actual biology of the PFC found in the brain. Our model also uses an unsupervised STDP learning rule to learn patterns and adapt to different task stimuli. Unlike previous studies, we have used real world dataset to study and record the behavior dynamics of the neurons. We show a novel way of simulating lesions by deactivating synapses. The effect of these lesions has been plotted as synaptic weights which gives a partial learning of the stimulus. Our experiments on task switching have been devised per research done on human subjects. The characteristic behaviors observed in our study helps in understanding and testing various hypotheses associated with neural structures involved in task switching. To the best of our knowledge, this model is the first to use STDP and a real dataset with the architecture of the PFC neurons derived from studies done on the brain using neuralimaging techniques.

\bibliographystyle{plain}
\bibliography{mybibliography}

\begin{thebibliography}{10}

\bibitem{bi1998synaptic}
Guo-qiang Bi and Mu-ming Poo.
\newblock Synaptic modifications in cultured hippocampal neurons: dependence on
  spike timing, synaptic strength, and postsynaptic cell type.
\newblock {\em Journal of neuroscience}, 18(24):10464--10472, 1998.

\bibitem{braver1995computational}
Todd~S Braver, Jonathan~D Cohen, and David Servan-Schreiber.
\newblock A computational model of prefrontal cortex function.
\newblock In {\em Advances in neural information processing systems}, pages
  141--148. Citeseer, 1995.

\bibitem{cohen1996computational}
Jonathan~D Cohen, Todd~S Braver, and Randall O{\'{}}~Reilly.
\newblock A computational approach to prefrontal cortex, cognitive control and
  schizophrenia: recent developments and current challenges.
\newblock {\em Philosophical transactions of the royal society of london.
  Series B: Biological sciences}, 351(1346):1515--1527, 1996.

\bibitem{diehl2015unsupervised}
Peter~U Diehl and Matthew Cook.
\newblock Unsupervised learning of digit recognition using
  spike-timing-dependent plasticity.
\newblock {\em Frontiers in computational neuroscience}, 9:99, 2015.

\bibitem{durstewitz2002computational}
Daniel Durstewitz and Jeremy~K Seamans.
\newblock The computational role of dopamine d1 receptors in working memory.
\newblock {\em Neural Networks}, 15(4-6):561--572, 2002.

\bibitem{durstewitz2000dopamine}
Daniel Durstewitz, Jeremy~K Seamans, and Terrence~J Sejnowski.
\newblock Dopamine-mediated stabilization of delay-period activity in a network
  model of prefrontal cortex.
\newblock {\em Journal of neurophysiology}, 2000.

\bibitem{funahashi1989mnemonic}
Shintaro Funahashi, Charles~J Bruce, and Patricia~S Goldman-Rakic.
\newblock Mnemonic coding of visual space in the monkey's dorsolateral
  prefrontal cortex.
\newblock {\em Journal of neurophysiology}, 61(2):331--349, 1989.

\bibitem{gerstner2002spiking}
Wulfram Gerstner and Werner~M Kistler.
\newblock {\em Spiking neuron models: Single neurons, populations, plasticity}.
\newblock Cambridge university press, 2002.

\bibitem{goard2016distinct}
Michael~J Goard, Gerald~N Pho, Jonathan Woodson, and Mriganka Sur.
\newblock Distinct roles of visual, parietal, and frontal motor cortices in
  memory-guided sensorimotor decisions.
\newblock {\em elife}, 5:e13764, 2016.

\bibitem{grant1948behavioral}
David~A Grant and Esta Berg.
\newblock A behavioral analysis of degree of reinforcement and ease of shifting
  to new responses in a weigl-type card-sorting problem.
\newblock {\em Journal of experimental psychology}, 38(4):404, 1948.

\bibitem{gupta2009hebbian}
Ankur Gupta and Lyle~N Long.
\newblock Hebbian learning with winner take all for spiking neural networks.
\newblock In {\em 2009 International Joint Conference on Neural Networks},
  pages 1054--1060. IEEE, 2009.

\bibitem{hertz2018introduction}
John~A Hertz.
\newblock {\em Introduction to the theory of neural computation}.
\newblock CRC Press, 2018.

\bibitem{lynch2019winner}
Nancy Lynch, Cameron Musco, and Merav Parter.
\newblock Winner-take-all computation in spiking neural networks.
\newblock {\em arXiv preprint arXiv:1904.12591}, 2019.

\bibitem{masquelier2008spike}
Timoth{\'e}e Masquelier, Rudy Guyonneau, and Simon~J Thorpe.
\newblock Spike timing dependent plasticity finds the start of repeating
  patterns in continuous spike trains.
\newblock {\em PloS one}, 3(1):e1377, 2008.

\bibitem{mehta2001neuronal}
Mayank~R Mehta.
\newblock Neuronal dynamics of predictive coding.
\newblock {\em The Neuroscientist}, 7(6):490--495, 2001.

\bibitem{moreno2014poisson}
Rub{\'e}n Moreno-Bote.
\newblock Poisson-like spiking in circuits with probabilistic synapses.
\newblock {\em PLoS computational biology}, 10(7):e1003522, 2014.

\bibitem{paugam2012computing}
H{\'e}lene Paugam-Moisy and Sander~M Bohte.
\newblock Computing with spiking neuron networks.
\newblock {\em Handbook of natural computing}, 1:1--47, 2012.

\bibitem{rainer1999prospective}
Gregor Rainer, S~Chenchal Rao, and Earl~K Miller.
\newblock Prospective coding for objects in primate prefrontal cortex.
\newblock {\em Journal of Neuroscience}, 19(13):5493--5505, 1999.

\bibitem{rao2002predictive}
R~Rao and T~Sejnowski.
\newblock Predictive coding, cortical feedback, and spike-timing dependent
  plasticity. probabilistic models of the brain, 297--315, 2002.

\bibitem{reggia1999disorders}
James~A Reggia, Eytan Ruppin, and DL~Glanzman.
\newblock {\em Disorders of brain, behavior, and cognition: the
  neurocomputational perspective}.
\newblock Elsevier, 1999.

\bibitem{riley2016role}
Mitchell~R Riley and Christos Constantinidis.
\newblock Role of prefrontal persistent activity in working memory.
\newblock {\em Frontiers in systems neuroscience}, 9:181, 2016.

\bibitem{rogers1995costs}
Robert~D Rogers and Stephen Monsell.
\newblock Costs of a predictible switch between simple cognitive tasks.
\newblock {\em Journal of experimental psychology: General}, 124(2):207, 1995.

\bibitem{rolls2007memory}
Edmund~T Rolls.
\newblock Memory, attention, and decision-making: A unifying computational
  neuroscience.
\newblock 2007.

\bibitem{rolls2008computational}
Edmund~T Rolls, Marco Loh, Gustavo Deco, and Georg Winterer.
\newblock Computational models of schizophrenia and dopamine modulation in the
  prefrontal cortex.
\newblock {\em Nature Reviews Neuroscience}, 9(9):696--709, 2008.

\bibitem{rubinstein1994task}
Joshua Rubinstein, Jeffrey~E Evans, and David~E Meyer.
\newblock Task switching in patients with prefrontal cortex damage.
\newblock In {\em Poster presented at the meeting of the Cognitive Neuroscience
  Society, San Francisco, CA}, 1994.

\bibitem{shallice2008multiple}
Tim Shallice, Donald~T Stuss, Terence~W Picton, Michael~P Alexander, and Susan
  Gillingham.
\newblock Multiple effects of prefrontal lesions on task-switching.
\newblock {\em Frontiers in human neuroscience}, 2:2, 2008.

\bibitem{song2000competitive}
Sen Song, Kenneth~D Miller, and Larry~F Abbott.
\newblock Competitive hebbian learning through spike-timing-dependent synaptic
  plasticity.
\newblock {\em Nature neuroscience}, 3(9):919--926, 2000.

\bibitem{stuss1984neuropsychological}
Donald~T Stuss and D~Frank Benson.
\newblock Neuropsychological studies of the frontal lobes.
\newblock {\em Psychological bulletin}, 95(1):3, 1984.

\bibitem{sudevan1987cuing}
Padmanabhan Sudevan and David~A Taylor.
\newblock The cuing and priming of cognitive operations.
\newblock {\em Journal of Experimental Psychology: Human perception and
  performance}, 13(1):89, 1987.

\bibitem{viswanathan2020study}
K~Ashwin Viswanathan, Goutam Mylavarapu, Kun Chen, and Johnson~P Thomas.
\newblock A study of prefrontal cortex task switching using spiking neural
  networks.
\newblock In {\em 2020 12th International Conference on Advanced Computational
  Intelligence (ICACI)}, pages 199--206. IEEE, 2020.

\bibitem{yeung2003effects}
Nick Yeung and Stephen Monsell.
\newblock The effects of recent practice on task switching.
\newblock {\em Journal of Experimental Psychology: Human Perception and
  Performance}, 29(5):919, 2003.

\end{thebibliography}

\end{document}